\pdfoutput=1

\documentclass[10pt,conference,compsocconf]{IEEEtran}
\usepackage{times}
\usepackage{multirow}
\usepackage{array}
\newcolumntype{L}[1]{>{\raggedright\let\newline\\\arraybackslash\hspace{0pt}}m{#1}}
\newcolumntype{C}[1]{>{\centering\let\newline\\\arraybackslash\hspace{0pt}}m{#1}}
\newcolumntype{R}[1]{>{\raggedleft\let\newline\\\arraybackslash\hspace{0pt}}m{#1}}
\usepackage{graphicx}
\usepackage{caption}
\usepackage[table,xcdraw]{xcolor}
\usepackage{pgfplots}
\usepackage{tikz}
\usepackage{xcolor}
\usepackage[utf8]{inputenc}
\usepackage{kotex}

\ifCLASSINFOpdf
\else
\fi
\usepackage{amsmath}
\begin{document}
\pagenumbering{gobble}
%
\title{\textbf{\Large CRAM: Clued Recurrent Aattentionttention Model }\\[0.2ex]}

\author{
\IEEEauthorblockN{Minki Chung, Sungzoon Cho\IEEEauthorrefmark{1}}
\IEEEauthorblockA{Data Mining Center, Department of Industrial Engineering,
Seoul National University, Seoul, Republic of Korea}
\IEEEauthorblockA{Email: minki.chung@dm.snu.ac.kr, zoon@snu.ac.kr}
}


\maketitle

\begin{abstract}
To overcome the poor scalability of convolutional neural network, recurrent attention model(RAM) selectively choose what and where to look on the image. By directing recurrent attention model how to look the image, RAM can be even more successful in that the given clue narrow down the scope of the possible focus zone. In this perspective, this work proposes clued recurrent attention model (CRAM) which add clue or constraint on the RAM better problem solving. CRAM follows encoder-decoder framework, encoder utilizes recurrent attention model with spatial transformer network and decoder which varies depending on the task. To ensure the performance, CRAM tackles two computer vision task. One is the image classification task, with clue given as the binary image saliency which indicates the approximate location of object. The other is the inpainting task, with clue given as binary mask which indicates the occluded part. In both tasks, CRAM shows better performance than existing methods showing the successful extension of RAM.

\end{abstract}



%
\IEEEpeerreviewmaketitle

\section{Introduction}
Adoption of convolutional neural network(CNN) \cite{lecun1989backpropagation} brought huge success on a lot of computer vision tasks such as classification and segmentation. One of limitation of CNN is its poor scalability with increasing input image size in terms of computation efficiency. With limited time and resources, it is necessary to be smart on selecting where, what, and how to look the image. Facing bird specific fine grained classification task, for example, it does not help much to pay attention on non-dog image part such as tree and sky. Rather, one should focus on regions which play decisive roles on classification such as beak or wings. If machine can learn how to pay attention on those regions will results better performance with lower energy usage. 

In this context, \textbf{Recurrent Attention Model(RAM)} \cite{mnih2014recurrent} introduces visual attention method in the problem of fine-grained classification task. By sequentially choosing where and what to look, RAM achieved better performance with lower usage of memory. Even more, attention mechanism tackled the vulnerable point, that deep learning model is the black box model by enabling interpretations of the results. But still there is more room for RAM for improvement. In addition to where and what to look, if one can give some clues on how to look, the task specific hint, learning could be more intuitive and efficient. From this insight, we propose the novel architecture, \textbf{Clued Recurrent Attention Model(CRAM)} which inserts problem solving oriented clue on RAM. These clues, or constraints give directions to machine for faster convergence and better performance. 

For evaluation, we perform experiments on two computer vision task  classification and inpainting. In classification task, clue is given as the binary saliency of the image which indicates the rough location of the object. In inpainting problem, clue is given as the binary mask which indicates the location the occluded region. Codes are implemented in tensorflow version 1.6.0 and uploaded at https://github.com/brekkanegg/cram.

In summary, the contributions of this work are as follows: 
\begin{enumerate}
  \item Proposed novel model clued recurrent attention model(CRAM) which inserted clue on RAM for more efficient problem solving.
  \item Defined clues for classification and inpainting task respectively for CRAM which are easy to interpret and approach.
  \item Evaluated on classification and inpainting task, showing the powerful extension of RAM. 
  \end{enumerate}

\section{Related Work}

\subsection{Reccurrent Attention Model(RAM)}

RAM \cite{mnih2014recurrent} first proposed recurrent neural network(RNN) \cite{mikolov2010recurrent} based attention model inspired by human visual system. When human are confronted with large image which is too big to be seen at a glance, he processes the image from part by part depending on his interest. By selectively choosing what and where to look RAM showed higher performance while reducing calculations and memory usage. However, since RAM attend the image region by using sampling method, it has fatal weakness of using REINFORCE, not back-propagation for optimization. After works of RAM, Deep Recurrent Attention Model(DRAM) \cite{ba2014multiple} showed advanced architecture for multiple object recognition and Deep Recurrent Attentive Writer(DRAW) \cite{gregor2015draw} introduced sequential image generation method without using REINFORCE. 

Spatial transformer network (STN) \cite{jaderberg2015spatial} first proposed a parametric spatial attention module for object classification task. This model includes a localization network that outputs the parameters for selecting region to attend in the input image. Recently, Recurrent Attentional Convolutional-Deconvolutional Network(RACDNN) \cite{kuen2016recurrent} gathered the strengths of both RAM and STN in saliency detection task. By replacing RAM locating module with STN, RACDNN can sequentially select where to attend on the image while still using back-propagation for optimization. This paper mainly adopted the RACDNN network with some technical twists to effectively insert the clue which acts as supervisor for problem solving.

\section{CRAM}

\begin{figure}[h]
    \centering
    \includegraphics[width=0.48\textwidth]{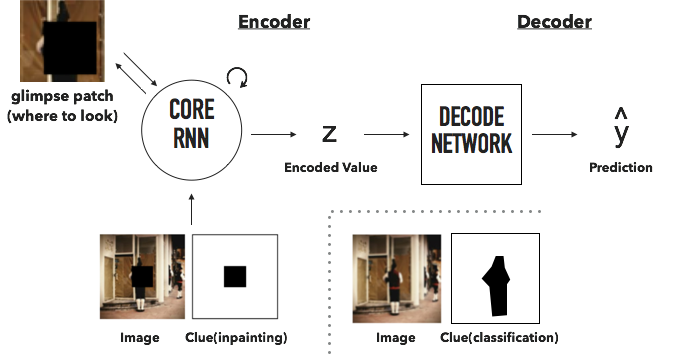}
    \caption{Overall architecture of CRAM. Note that image and clue become different depending on the task (left under and right under).}
    \label{fig:overall}
\end{figure}

The architecture of CRAM is based on encoder-decoder structure. Encoder is similar to RACDNN\cite{kuen2016recurrent} with modified spatial transformer network\cite{jaderberg2015spatial} and inserted clue. While encoder is identical regardless of the type of task, decoder becomes different where the given task is classification or inpainting. Figure \ref{fig:overall} shows the overall architecture of CRAM. 

\subsection{\bf{Encoder}}
\begin{figure}[h]
    \centering
    \includegraphics[width=0.48\textwidth]{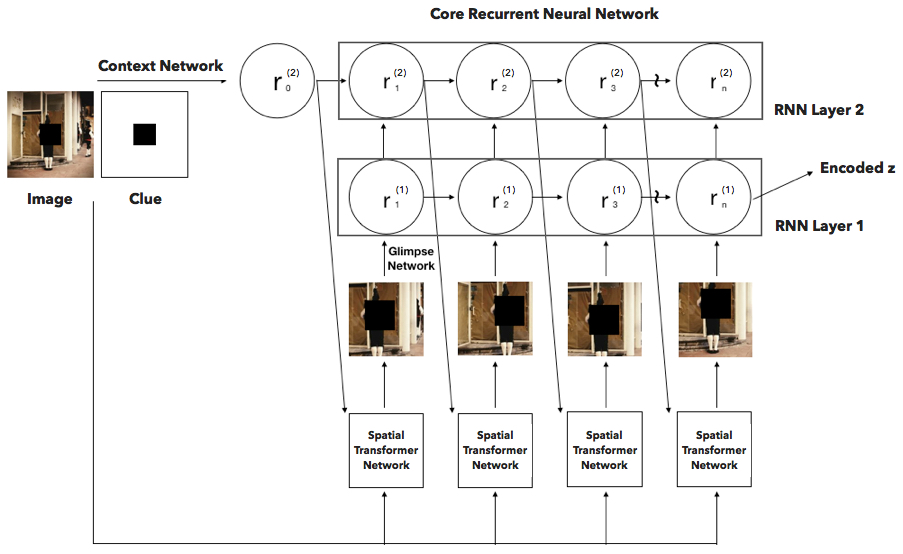}
    \caption{Architecture of CRAM encoder. Note that the image is for inpainting task, where clue is given as binary mask that indicates the occluded region.}
    \label{fig:enc}
\end{figure}

Encoder is composed of 4 subnetworks: context network, spatial transformer network, glimpse network and core recurrent neural network. The overall architecture of encoder is shown in Figure \ref{fig:enc}. Considering the flow of information, we will go into details of each network one by one.

\textbf{Context Network: }The context network is the first part of encoder which receives image and clue as inputs and outputs the initial state tuple of {$r_{0}^{2}$}. {$r_{0}^{2}$} is first input of second layer of core recurrent neural network as shown in Figure \ref{fig:enc}. Using downsampled image{$(i_{coarse}),$} and downsampled clue{$(c_{coarse})$}, context network provides reasonable starting point for choosing image region to concentrate. Downsampled image and clue are processed with CNN followed by MLP respectively.  

\begin{align}\label{eq:cn}
c_{0} = MLP_{c}(CNN_{context}(i_{coarse}, c_{coarse})) \\
h_{0} = MLP_{h}(CNN_{context}(i_{coarse}, c_{coarse})) 
\end{align}
where ({$c_{0}$}, {$h_{0}$}) is the first state tuple of {$r_{0}^{2}$}.  

\textbf{Spatial Transformer Network: } Spatial transformer network(STN) select region to attend considering given task and clue \cite{jaderberg2015spatial}. Different from existing STN, CRAM uses modified STN which receives image, clue, and output of second layer of core RNN as an inputs and outputs glimpse patch. From now on, glimpse patch indicates the attended image which is cropped and zoomed in. Here, the STN is composed of two parts. One is the localization part which calculates the transformation matrix {$\tau$} with CNN and MLP. The other is the transformer part which zoom in the image using the transformation matrix {$\tau$} above and obtain the glimpse. The affine transformation matrix {$\tau$} with isotropic scaling and translation is given as Equation \ref{eq:tau}. 

\begin{equation}\label{eq:tau}
\tau = \begin{bmatrix}
s & 0 & t_{x} \\ 
0 & s & t_{y}\\ 
0 & 0 & 1
\end{bmatrix}
\end{equation}
where {$s, t_{x}, t_{y}$} are the scaling, horizontal translation and vertical translation parameter respectively.

\begin{figure}[h]
    \centering
    \includegraphics[width=0.48\textwidth]{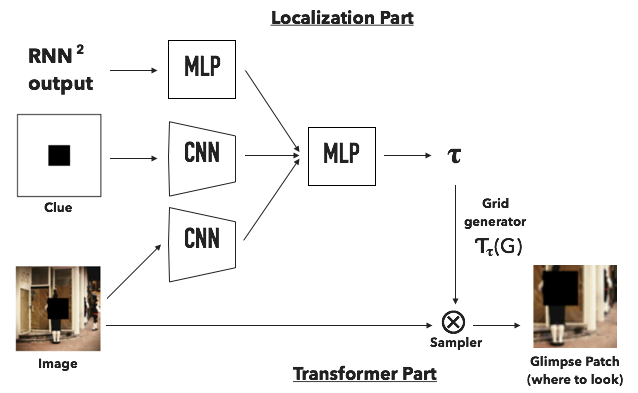}
    \caption{Architecture of STN. STN is consists of localisation part which calculates {$\tau$} and transformer part which obtain glimpse. }
    \label{fig:stn}
\end{figure}

Total process of STN is shown in Figure \ref{fig:stn}.

In equations, the process STN is as follows:

\begin{equation}\label{eq:sn}
glimpse\_patch_{n} = STN(image, clue, \tau_{n})
\end{equation}
where {$n$} in {$ glimpse\_patch_{n}$} is the step of core RNN ranging from 1 to total glimpse number. {$\tau$} is obtained by the below equation.

\begin{equation}\label{eq:en}
\tau_{n} = MLP_{loc}(CNN_{i}(image)\oplus CNN_{c}(clue)\oplus MLP_{r}(r_{n}^{(2)}))
\end{equation}
where {$\oplus$} is concat operation.

\textbf{Glimpse Network: }The glimpse network is a non-linear function which receives current glimpse patch, {$ glimpse\_patch_{n}(gp_{n}$)} and attend region information, {$\tau$} as inputs and outputs current step glimpse vector. Glimpse vector is later used as the input of first core RNN. {$glimpse\_vector_{n}(gv_{n})$} is obtained by multiplicative interaction between extracted features of {$glimpse\_patch_{n}$} and {$\tau$}. The method of interaction is first proposed by \cite{larochelle2010learning}. Similar to other mentioned networks, CNN and MLP are used for feature extraction. 

\begin{equation}\label{eq:gn}
\begin{split}
gv_{n} = MLP_{what}(CNN_{what}(gp_{n})) \odot MLP_{where}(\tau_{n})
\end{split}
\end{equation}

where {$\odot$} is a element-wise vector multiplication operation. 

\textbf{Core Recurrent Neural Network: } Recurrent neural network is the core structure of CRAM, which aggregates information extracted from the stepwise glimpses and calculates encoded vector z. Iterating for set RNN steps(total glimpse numbers), core RNN receives {$glimpse\_vector_{n}$} at the first layer. The output of second layer {$r_{n}^{(2)}$} is again used by spatial transformer network's localization part as Equation \ref{eq:en}. 

\begin{equation}\label{eq:rn}
r_{n}^{(1)} = R_{recur}^{ 1}(glimpse\_vector_{n}, r_{n-1}^{(1)}) \\
\end{equation}
\begin{equation}\label{eq:rn}
r_{n}^{(2)} = R_{recur}^{ 2}(r_{n}^{(1)}, r_{n-1}^{(2)})
\end{equation}

\subsection{\bf{Decoder}}

\subsubsection{Classification}
\begin{figure}[h]
    \centering
    \includegraphics[width=0.48\textwidth]{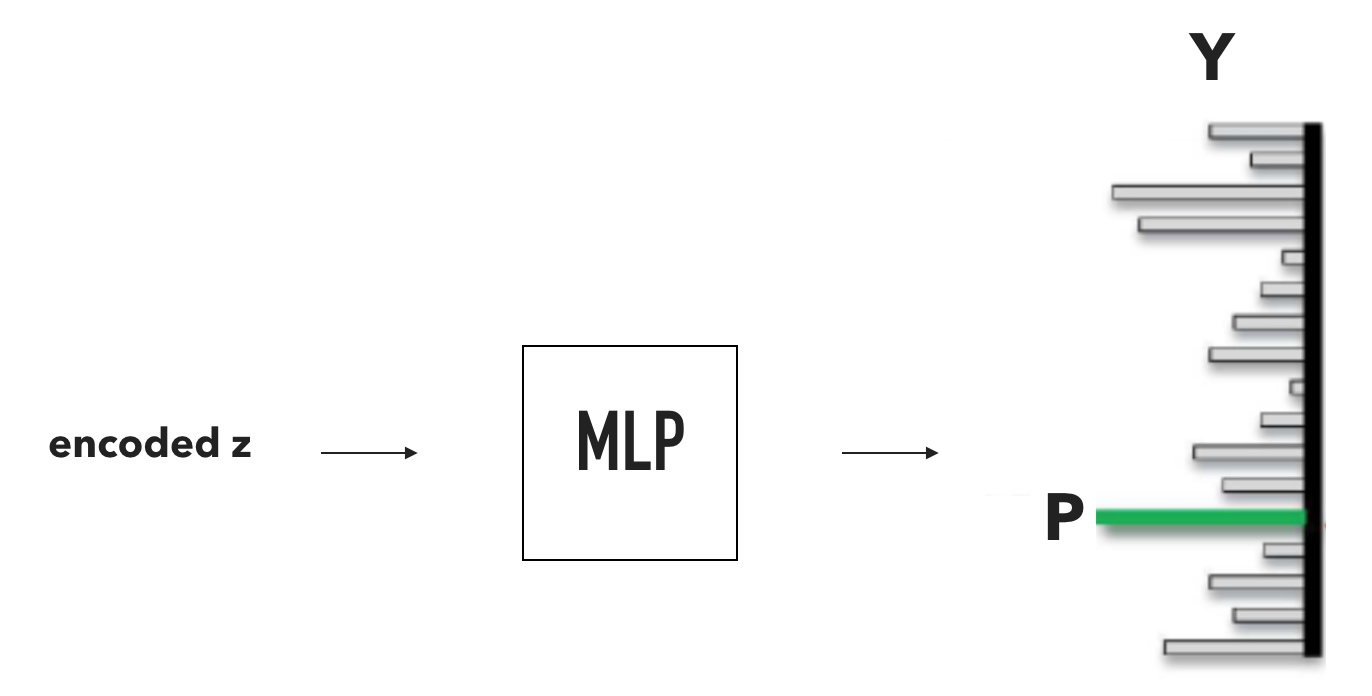}
    \caption{Architecture of CRAM decoder for image classification.}
    \label{fig:deccls}
\end{figure}

Like general image classification approach, encoded z is passed through MLP and outputs possibility of each class. Image of decoder for classification is shown in Figure \ref{fig:deccls}.

\subsubsection{Inpainting}
\begin{figure}[h]
    \centering
    \includegraphics[width=0.5\textwidth]{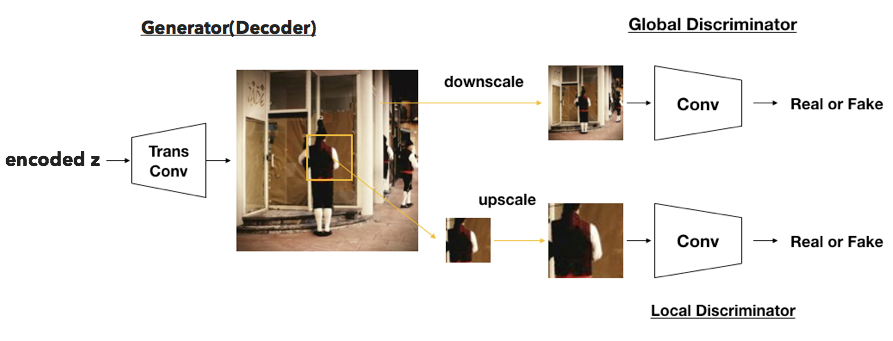}
    \caption{Architecture of CRAM decoder and discriminators for image inpainting.}
    \label{fig:dec}
\end{figure}

Utilizing the architecture of DCGAN \cite {radford2015unsupervised}, contaminated image is completed starting from the the encoded z from the encoder. To ensure the quality of completed image, we adopted generative adversarial network(GAN)\cite{goodfellow2014generative} framework in both local and global scale \cite{iizuka2017globally}. Here decoder works as generator and local and global discriminators evaluate its plausibility in local and global scale respectively. Image of decoder for inpainting is shown in Figure \ref{fig:dec}.

\section{Training}
Loss function of CRAM can be divided into two: encoder related loss({$L_{enc}$}) and decoder-related loss({$L_{dec}$}). {$L_{enc}$} constraints the glimpse patch to be consistent with the clue. For classification task, where the  clue is object saliency, it is favorable if the glimpse patches covers the salient part at most. For inpainting case, there should be a supervisor that urges the glimpse patches contains the occluded region considering the region neighbor of occlusion are the most relevant part for completion. In order to satisfy above condition for both classification and inpainting cases {$L_{enc}$} or {$L_{clue}$} is as follows:

\begin{equation}\label{eq:lossg}
L_{enc}=L_{clue}(clue, STN, \tau) = \sum_{n}{ST(clue, \tau_{n})} 
\end{equation}

where {$STN$} is the trained spatial transformer network Equation \ref{eq:sn} and {$\tau_{n}$} is obtained from Equation \ref{eq:tau} in each step of core RNN. Decoder loss which is different depending on the given task will be dealt separately shortly. Note that clue is binary image for both classification and inpainting tasks. 

Since {$L_{dec}$} is different depending on whether the problem is classification or completion, further explanations for losses will be divided into two.   

\subsection{Classification}
Decoder related loss in image classification task utilize cross entropy loss like general classification approach. Then total loss {$L_{tot-cls}$} for image classification becomes:
 
\begin{align}\label{eq:losscls}
L_{tot-cls} &= L_{enc} + L_{dec} \\
& = L_{clue}(clue, ST, \tau s) + L_{cls}(Y, Y^{*}) 
\end{align}

where clue is the binary image which takes the value of 1 for salient part and otherwise 0, and {$Y$} and {$Y^{*}$} are predicted class label vector and ground truth class label vector respectively.

\subsection{Inpainting}
Decoder related loss for image inpainting consists of reconstruction loss and gan loss. 

Reconstruction loss helps completion to be more stable and gan loss enables better quality of restoration. For reconstruction loss L1 loss considering contaminated region of input is used:

\begin{equation}\label{eq:reconloss}
L_{recon}(z, clue, Y^{*}) = \| clue \odot (G(z) - Y^{*}) \| _{1}  
\end{equation}
where z is encoded vector from the encoder, clue is the binary image which takes the value of 1 for occluded region and otherwise 0, G is generator(or decoder) and {$Y^{*}$} is the original image before contamination.

Since there are two discriminators, local and global scale gan loss is summation of local gan loss and global gan loss.
\begin{equation}\label{eq:ganlosses}
\begin{split}
L_{gan} &= L_{global\_gan} + L_{local\_gan}
\end{split}
\end{equation}

GAN loss for local and global scale are defined as follows: 
\begin{equation}\label{eq:ganloss}
\begin{split}
L_{local\_gan} &= log(1-D_{local}(Y^{*} \odot clue)) \\ &+ logD_{local}(G(image, clue) \odot clue) \\
\end{split}
\end{equation}

\begin{equation}\label{eq:ganloss2}
\begin{split}
L_{global\_gan} &= log(1-D_{global}(Y^{*} ))\\ &+ logD_{global}(G(image, clue))
\end{split}
\end{equation}

Combining Equation \ref{eq:lossg}, \ref{eq:reconloss} and \ref{eq:ganlosses}, the total loss for image inpainting {$L_{tot-ip}$} becomes:

\begin{align}\label{eq:ganloss3}
L_{tot-ip} &= L_{enc} + L_{dec} \\
&= L_{clue} + \alpha L_{recon} +\beta L_{gan}
\end{align}
where {$\alpha$} and {$\beta$} is weighting hyperparameter and {$L_{gan}$} is summation of {$L_{global\_gan}$} and {$L_{global\_gan}$}.

\section{Implementation Details}

\subsection{Classification}
\begin{figure}[h]
    \centering
    \includegraphics[width=0.48\textwidth]{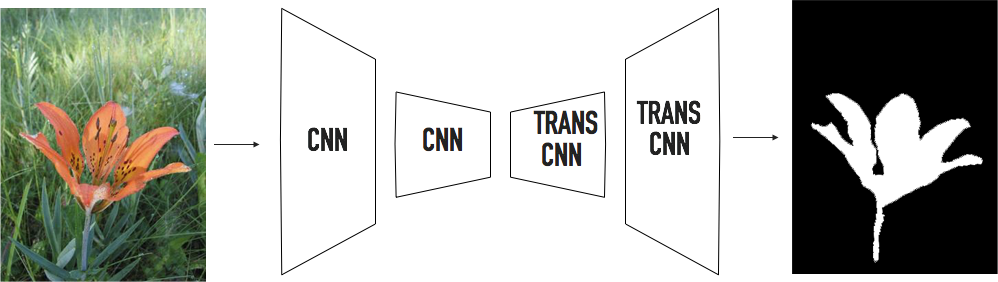}
    \caption{CNN-DecNN to obtain rough saliency of image. This rough saliency is the clue classification task.}
    \label{fig:cnndecnn}
\end{figure}

In order to obtain the clue, saliency map, we use a convolutional deconvolutional network (CNN-DecNN) \cite{noh2015learning} as shown in Figure \ref{fig:cnndecnn}. CNN-DecNN is pre-trained with the MSRA10k\cite{cheng2015global} dataset, which is by far the largest publicly available saliency detection dataset, containing 10,000 annotated saliency images. This CNN-DecNN is trained with Adam\cite{kingma2014adam} in default learning settings. In training and inference period, rough saliency(or clue) is obtained from the pre-trained CNN-DecNN. 

As mentioned earlier, encoder consists of 4 subnetworks: context network, spatial transformer network, glimpse network, and core RNN.
Image and clue is 4 times downsampled and used an input of context network. Each passes 3 layer-CNN(3 x 3 kernel size, 1 x 1 stride, same zero padding) each followed by max-pooling layer(3 x 3 kernel size, 2 x 2 stride, same zero padding) and outputs vectors. These vectors are concatenated and once again passes 2 layer MLP and outputs the initial state for second layer of core RNN. 
Localization part of spatial transformer network consists of CNN and MLP. For image and clue input, 3 layer-CNN(5 x 5 kernel size, 2 x 2 stride, same zero padding) is applied. 2 layer-MLP is applied in second core RNN output input. Output vectors of CNN and MLP are concatenated and once again pass through 2-layer MLP for {$s, t_{x}, t_{y}$}, the parameters of {$\tau$}.
Glimpse network receives glimpse patch and {$\tau$} above as an input. 1-layer MLP is applied on {$\tau$} while Glimpse patch passes through 3-layer CNN and 1-layer MLP to match the vector length with the {$\tau$} vector after 1-layer MLP. Glimpse vector is obtained by element-wise vector multiplication operation of above output vectors. 
Core RNN is composed of 2 layer with Long-Short-Term Memory (LSTM) units for \cite{hochreiter1997long} for of its ability to learn long-range dependencies and stable learning dynamics.  
Decoder is quite simple, only made up of 3-layer MLP.
Filter number of CNNs, dimensions of MLPs, dimensions of core RNNs, number of core RNN steps vary depending on the size of the image.
All CNN and MLP except last layer includes batch normalization\cite{ioffe2015batch} and elu activation\cite{clevert2015fast}.
We used Adam optimizer \cite{kingma2014adam} with learning rate 1e-4. 

\subsection{Inpainting}
Encoder settings are identical with image classification case.
Decoder(or generator) consists of fractionally-strided CNN(3 x 3 kernel size, 1/2 stride) until the original image size are recovered. 
Both local and global discriminators are based on CNN, extracts the features from the image to judge the input genuineness. Local discriminator is composed of 4 layer-CNN(5 x 5 kernel size, 2 x 2 stride, same zero padding) and 2-layer MLP. Global discriminator consists of 3-layer CNN(5 x 5 kernel size, 2 x 2 stride, same zero padding)and  2-layer MLP. Sigmoid function is applied on the last outputs of local and global discriminator in order to ensure the output value to be between 0 and 1. All CNN, fractionally-strided CNN, and MLP except last layer includes batch normalization and elu activation. Same as classification settings, filter number of CNNs, filter number of fractionally-strided CNNs, dimensions of MLPs, dimensions of core RNNs, number of core RNN steps vary depending on the size of the image.

\section{Experiment}

\subsection{Image Classification}
Work in progress.

\subsection{Image Inpainting}
\subsubsection{Dataset}
Street View House Numbers (SVHN) dataset\cite{netzer2011reading} is a real world image dataset for object recognition obtained from house numbers in Google street view image. SVHN dataset contains 73257 training digits and 26032 testing digits size of 32 x 32 in RGB color scale.  

\subsubsection{Result}
\begin{figure}[h]
    \centering
    \includegraphics[width=0.4\textwidth]{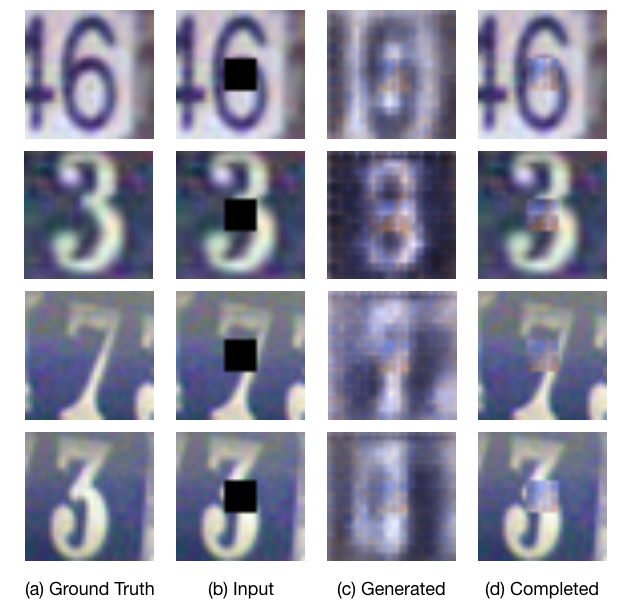}
    \caption{Experiment results on SVHN. From left to right is ground truth, input contaminated image, generated image by CRAM decoder and finally the completed image which was partially replaced with generated image only for missing region.}
    \label{fig:svhn}
\end{figure}

Figure \ref{fig:svhn} showed the result of inpainting with SVHN dataset. 6.25\% pixels of image at the center are occluded. Even though the result is not excellent, it is enough to show the possibility and scalability of CRAM. With better generative model, it is expected to show better performance.

\section{Conclusion}
Work in progress.





%
%
%

\bibliographystyle{IEEEtran}
\bibliography{egbib}

\clearpage

\end{document}